\documentclass[conference]{IEEEtran}

\usepackage[T1]{fontenc}
\usepackage[utf8]{inputenc}
\usepackage{amsmath,amssymb}
\usepackage{graphicx}
\usepackage{booktabs}
\usepackage{array}
\usepackage{cite}
\usepackage{url}
\usepackage{hyperref}
\usepackage{xcolor}

\hypersetup{
    colorlinks=true,
    linkcolor=black,
    citecolor=black,
    urlcolor=blue
}
\hypersetup{
    pdftitle={An Exploratory Ablation of a Small MLA--SSM Hybrid Language Model},
    pdfauthor={Christos Koutsiaris},
    pdfsubject={Single-seed ablation study of a small hybrid language model},
    pdfkeywords={language models, state space models, multi-head latent attention, ablation study}
}

\title{An Exploratory Ablation of a Small MLA--SSM\\Hybrid Language Model}

\author{
  \IEEEauthorblockN{Christos Koutsiaris}
  \IEEEauthorblockA{Development Expert, SAP P\&E, Cloud ERP, UX Foundation\\
  christos.koutsiaris@sap.com\\
  Code and data: \url{https://github.com/unseen1980/talh}}
}

\begin{document}
\maketitle

\begin{abstract}
We report an exploratory, single-seed ablation of TALH (Adaptive Latent Hybrid),
a decoder-only language model with parallel Multi-head Latent Attention (MLA)
and a custom recurrent state-space (SSM) branch. Five variants, spanning
117--217M estimated active parameters per token, are trained from scratch on a
FineWeb sample for the same number of optimisation steps and tokens. In this
specific setup, removing the SSM branch gives the largest degradation in
validation perplexity (MLA-only PPL~315), whereas removing MLA has a much smaller
effect (SSM-only PPL~239). A dense-FFN hybrid obtains PPL~231, compared with 240
for the tested top-2 ternary-MoE hybrid, while using 3.87~GB less peak training
memory. We also preserve a preliminary Apple M3 timing observation: among the
five unoptimised implementations, MLA-only has the flattest measured
time-to-first-token curve from 512 to 2{,}048 prompt tokens, although the dense
Transformer is much faster in absolute terms. Because the runs are single-seed,
parameter counts are unmatched, the evaluation stream may overlap the training
source, and raw repeated timing records are unavailable, these results support
implementation-specific hypotheses rather than general conclusions about MLA,
SSMs, or mixture-of-experts models.
\end{abstract}


\section{Introduction}

Running capable language models locally, without sending data to a remote
server, is a practical goal that standard Transformer architectures make
difficult. Two bottlenecks dominate. First, the key-value (KV) attention
cache grows linearly with context length, attention prefill is quadratic in
sequence length, and each incremental decode step attends over the accumulated
cache. Second, even at moderate context lengths, the total memory
footprint of a competitive model often exceeds what commodity hardware provides.

Several architectural directions have been proposed to address one or both
bottlenecks. Grouped-query and multi-query attention reduce KV cache size by
sharing keys and values across heads~\cite{gqa2023}. State space models
(SSMs) such as Mamba replace attention entirely with a recurrent state update
that has constant memory in context length~\cite{mamba2023}. Hybrid
architectures interleave or combine attention and SSM layers to get the
complementary strengths of both~\cite{h3_2022,jamba2024,samba2024}. More
recently, DeepSeek-V2 introduced Multi-head Latent Attention (MLA), which
compresses the KV cache via a low-rank joint factorisation of keys and
values~\cite{deepseekv2}, and Hymba proposed parallel within-layer
attention+SSM hybrid heads specifically targeting small, on-device
models~\cite{hymba2024}. Zebra-Llama explicitly combined MLA with SSM layers
and reported large KV-cache reductions at 1--8B scale~\cite{zebra2025}.
Subsequent systematic experiments also show that hybrid conclusions depend on
component allocation, fusion, and placement~\cite{hybrids2026}.

TALH sits in this same design space: it combines MLA with a parallel recurrent
branch and optional sparse expert routing. Rather than claiming a novel
architecture, we report an \textbf{exploratory ablation study} of one small
implementation and ask how its components behave under a constrained training
budget. Concretely:

\begin{itemize}
  \item Which component, MLA or SSM, is the primary quality driver in the hybrid?
  \item Does MLA's KV-cache compression translate into measurably better
        inference latency scaling on real Apple Silicon hardware?
  \item Does ternary MoE routing improve over a plain dense feed-forward layer
        at this scale?
  \item Under the same step and token budget, how does the hybrid compare
        against a smaller standard Transformer baseline?
\end{itemize}

We investigate these questions with a five-way, single-seed ablation trained on
FineWeb, followed by preliminary measurements on a MacBook M3. Phase~1 uses the
same training steps, nominal tokens, data stream, and curriculum for every
variant. It is not FLOP-, parameter-, or wall-clock-matched, so the results are
descriptive rather than estimates of an architecture-only causal effect.

\section{Related Work}

\textbf{KV-cache reduction.} Grouped-query attention (GQA) and multi-query
attention (MQA) reduce cache size by sharing K/V projections across heads and
are now standard in production models~\cite{gqa2023}. DeepSeek-V2's MLA
takes a different approach: a low-rank joint factorisation that compresses the
full KV state into a small latent vector, claiming a 93\% cache reduction while
maintaining quality at 236B total parameters~\cite{deepseekv2}. TALH tests MLA
in variants spanning approximately 117--217M active parameters and records a
preliminary latency observation.

\textbf{State space models.} S4 and its successors established that SSMs can
model long sequences with linear memory~\cite{mamba2023,h3_2022}. Mamba
introduced selective SSMs with hardware-aware algorithms competitive in
language modelling with Transformers~\cite{mamba2023}. H3 showed that a
hybrid SSM+attention model can outperform pure Transformers in perplexity on
OpenWebText and scale to the Pile with faster inference~\cite{h3_2022}.

\textbf{Hybrid attention-SSM architectures.} Jamba interleaves Mamba and
Transformer layers with MoE at scale~\cite{jamba2024}. Samba combines Mamba
with sliding-window attention for long-context efficiency~\cite{samba2024}.
Hymba is the closest prior work to our design: it proposes a \emph{parallel}
hybrid-head layer integrating attention heads and SSM heads within the same
layer for small/on-device models, and reports large cache reductions and
throughput gains~\cite{hymba2024}. Zebra-Llama explicitly combines MLA and
Mamba-2 layers via a post-training pipeline, achieving dramatic KV-cache
compression at 1--8B scale while retaining benchmark performance~\cite{zebra2025}.

\textbf{Positioning.} Compared with Hymba and Zebra-Llama, TALH differs in
three ways: (1) we train from scratch rather than fine-tuning an existing
checkpoint; (2) we explicitly ablate MoE routing vs.\ dense FFN within the
hybrid; and (3) we include a preliminary Apple Silicon timing observation. Our
contribution is the transparent ablation and negative result at a smaller scale,
with the evidential limitations described in Section~VII.

\textbf{MoE scaling.} Sparse expert routing has shown large efficiency gains
at model scales of tens to hundreds of billions of
parameters~\cite{deepseekv2}. Whether a particular sparse configuration helps
at smaller scales depends on routing, expert capacity, precision, and the
comparison budget; our experiment records one negative result rather than a
general scale threshold. In particular, compact MoEs have shown gains around
the 200M-active-parameter regime under different designs and controls
~\cite{cosmoe2025}.

\section{Architecture}
\label{sec:arch}

TALH is built from three architectural components.

\subsection{Multi-head Latent Attention (MLA)}

MLA~\cite{deepseekv2} compresses the KV state into a low-dimensional latent
vector before storing it in the cache. At inference time the latent is
projected back up into the full key and value space. The cache stores the
small latent rather than the full KV vectors, so cache size grows with the
latent dimension (200 in our setup) rather than with $d_{\text{model}}$ (800),
giving a theoretical compression ratio of $800/200 = 4\times$ per head. In
our preliminary 15M experiments we measured an $8\times$ reduction in total
cache memory relative to standard multi-head attention.

\subsection{State Space Model (SSM) Branch}

The tested branch is a custom minimal selective recurrence, not Mamba or
Mamba-2. It maintains a fixed-size hidden state and evaluates a sequential
update for each token; its state is constant in sequence length and its
arithmetic work is linear in sequence length. The implementation is intended
for architectural exploration rather than as a production SSM kernel. In TALH,
the recurrent branch runs in parallel with MLA; a learned projection followed
by an elementwise sigmoid gate combines their outputs.

\subsection{Ternary Mixture-of-Experts (MoE)}

In the \texttt{full} variant, the feed-forward block uses ternary-weighted
expert networks inspired by BitNet~\cite{bitnet2025}. Each forward pass
quantises an FP32 master weight to $\{-1,0,+1\}$ using a straight-through
estimator; the training implementation does not pack the master weights into a
1.58-bit representation. At each token, a router selects the top-2 of 8
experts. The lowest-loss variant in this ablation,
\texttt{dense\_ffn}, \emph{does not} use ternary weights; it uses a standard
BF16 SwiGLU feed-forward layer. In this run, the tested ternary-MoE
configuration provides no validation-loss benefit. The \texttt{full}
variant is included as an ablation target, not as the recommended design.

\subsection{Ablation Variants and Parameter Counts}

Table~\ref{tab:variants} lists total parameters and an estimated number touched
per token. For MoE variants, the estimate counts the embedding and all
non-expert parameters plus two of eight experts in every layer. It is a
parameter-access convention, not a FLOP estimate: shared weights operate over
all sequence positions, and the custom recurrent branch has substantially
different execution cost from attention.

\begin{table}[!t]
\centering
\caption{Architectural variants ($d_\text{model}=800$, $d_\text{ff}=1{,}600$,
         12~layers, 8 experts, top-$k=2$). Total = all stored weights;
         Active = weights touched per forward token pass.}
\label{tab:variants}
\resizebox{\columnwidth}{!}{%
\begin{tabular}{lp{3.2cm}cc}
\toprule
Variant & Components & Total & Active \\
\midrule
\texttt{full}      & MLA+SSM+ternary MoE & 493M & $\approx$217M \\
\texttt{ssm\_only} & SSM+MoE (no MLA)    & 480M & $\approx$203M \\
\texttt{mla\_only} & MLA+MoE (no SSM)    & 453M & $\approx$177M \\
\texttt{dense\_ffn}& MLA+SSM+dense FFN   & 171M & 171M \\
\texttt{transformer\_dense} & MHA+dense FFN & 117M & 117M \\
\bottomrule
\end{tabular}}
\end{table}

\textbf{Note on FFN width.} All variants use $d_\text{ff}=1{,}600$, a
2$\times$ ratio to the hidden width of 800. Standard practice often uses a
4$\times$ ratio ($d_\text{ff}=3{,}200$). The same width is used throughout,
but it need not affect different architectures equally. A wider FFN for the baseline would increase its parameter
count toward 163M and may reduce the quality gap; this is left for future work.

\section{Experimental Setup}

\subsection{Preliminary Experiments at 15M Parameters}

\begin{figure}[!t]
  \centering
  \includegraphics[width=\linewidth]{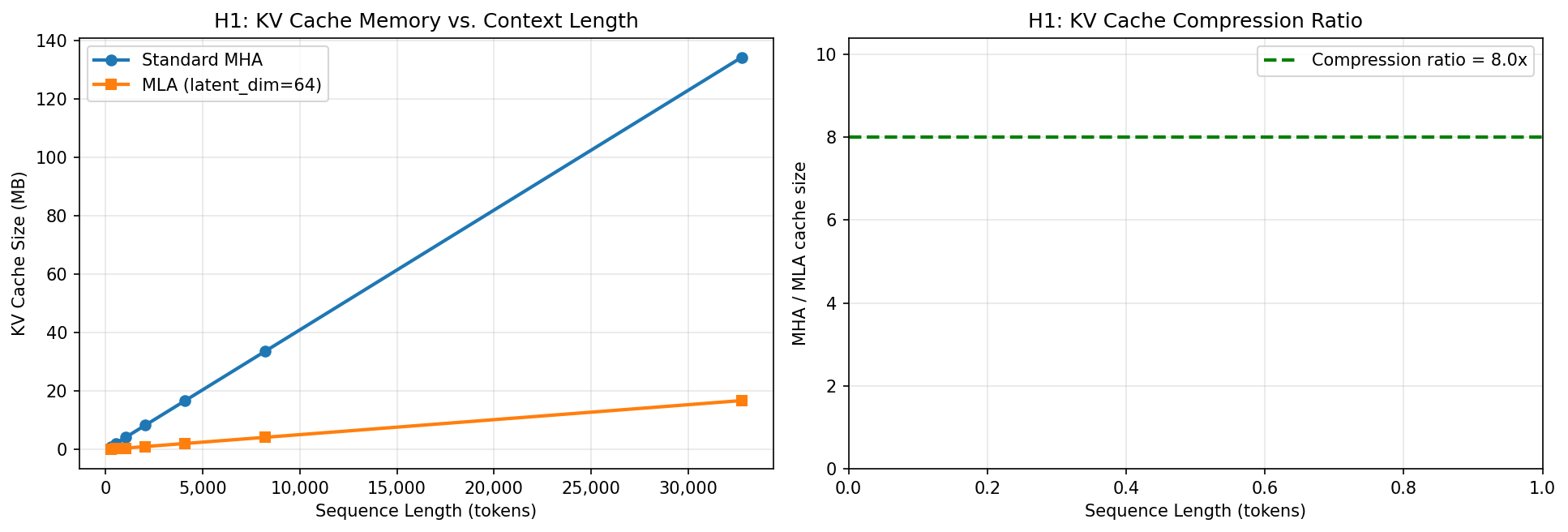}
  \caption{15M pilot: MLA reduces KV-cache size by $8\times$ relative to
           standard multi-head attention.}
  \label{fig:kvcache}
\end{figure}

\begin{figure}[!t]
  \centering
  \includegraphics[width=\linewidth]{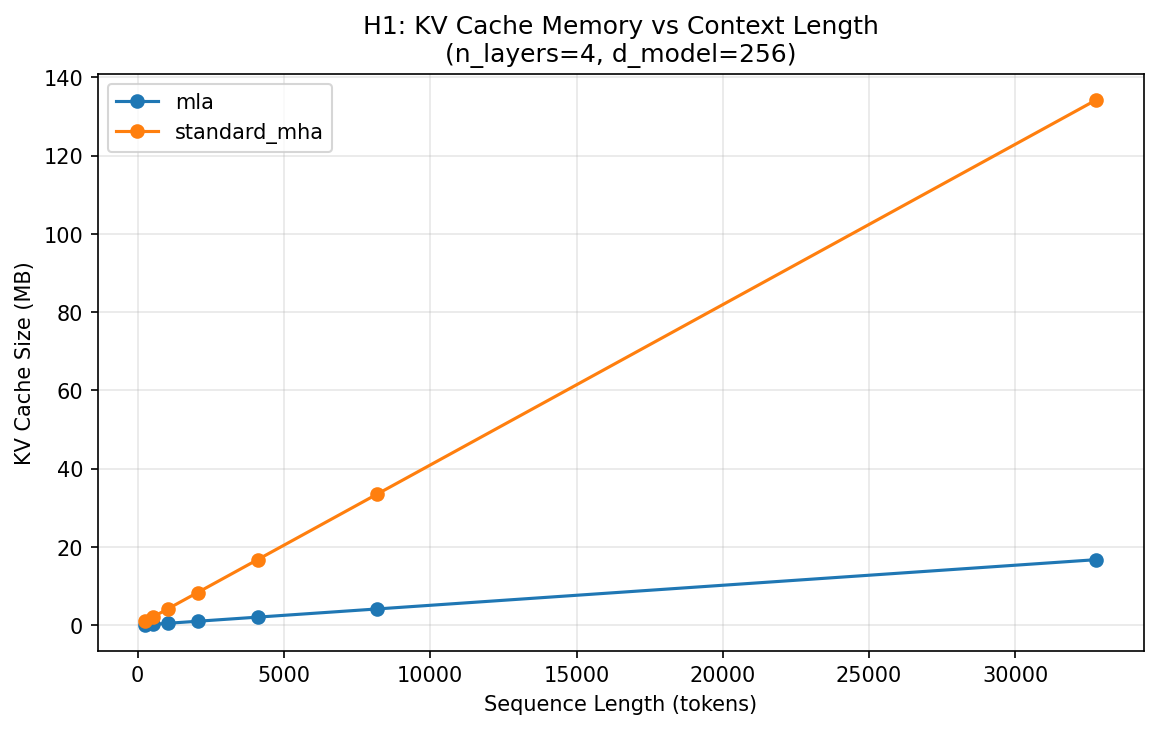}
  \caption{15M pilot: total memory on Apple Silicon stays within the 18~GB
           budget up to 4{,}096-token context.}
  \label{fig:memory}
\end{figure}

Before the main study we run a smaller set of experiments at approximately 15M
active parameters on the MacBook M3 to confirm local runnability and measure
KV-cache compression empirically. Figs.~\ref{fig:kvcache}
and~\ref{fig:memory} show the results. MLA achieves an $8\times$ cache
reduction relative to standard MHA, and peak memory grows from 146~MB at 256
tokens to 832~MB at 4{,}096 tokens, within the 18~GB unified-memory budget.
At this scale, \texttt{full} and \texttt{dense\_ffn} achieved validation
perplexities of 1.93 and 1.49 against 3.00 for a GPT-Neo reference.

\subsection{Main Training on FineWeb}
\label{sec:setup_main}

\textbf{Dataset.} We train on two local parquet shards from the FineWeb
\texttt{sample-10BT} subset~\cite{fineweb2024} using the GPT-2 tokeniser
(50{,}257-token vocabulary). The local-data evaluation path streams from the
same shard collection after skipping 10{,}000 documents. This creates a
deterministic evaluation stream, but the released logs do not establish that
it is disjoint from every document consumed during training. We therefore use
``validation'' as the implementation's label rather than claiming a clean
held-out estimate. Code, configs, training logs, and results are available at
\url{https://github.com/unseen1980/talh}.

\textbf{Hardware.} All training runs use a single NVIDIA A100-SXM4-40GB GPU
(Vast.ai, 1 GPU instance).

\textbf{Context-length curriculum.} We use a staged curriculum:
steps~0--2{,}999 use 256-token sequences; steps~3{,}000--4{,}999 use
512-token sequences; steps~5{,}000--7{,}000 use 1{,}024-token sequences.

\textbf{Phase~1: budget scan (all five variants).} All five variants are
trained for 4{,}000 steps (covering the 256-token and 512-token stages). Every
variant receives the same number of optimisation steps and nominal tokens,
with the same curriculum and source stream. Parameter counts, operations, and
wall-clock time are not matched. Phase~1 is the primary comparison point.

\textbf{Phase~2: extended training (\texttt{full} and \texttt{dense\_ffn}
only).} Based on Phase~1 results, these two hybrid variants are
extended to 7{,}000 steps, adding the 1{,}024-token curriculum stage. The
remaining three variants are not extended due to GPU budget constraints.
Phase~2 results are reported separately and cannot be directly compared
against Phase~1-only variants.

\textbf{Optimiser.} AdamW, cosine decay, initial LR $10^{-4}$, 400 warm-up
steps, gradient clipping at 0.5. BF16 mixed precision.

\textbf{Evaluation.} Validation perplexity is computed at sequence lengths 512
and 1{,}024 on the evaluation stream described above. There is one training
run per variant and no multi-seed replication.

\subsection{Local Inference Benchmark}

Best checkpoints were loaded on a MacBook M3 (18~GB unified memory) using
prototype conversion utilities. We recorded time-to-first-token (TTFT), defined as the wall-clock time from
prompt submission to the first generated token, at three prompt lengths:
512, 1{,}024, and 2{,}048 tokens. Decode throughput
(tokens/s) is not reported. All runs used an unoptimised
prototype stack without quantisation, kernel fusion, or prefix caching;
absolute latency numbers should not be compared directly against
production-optimised inference frameworks. The original checkpoints and raw
per-trial timing records are not included in the repository, so these values
are retained as preliminary observations and are not independently
reproducible from the release.

\section{Results}

\subsection{Step- and Token-Matched Quality: Phase~1}
\label{sec:phase1}

\begin{figure}[!t]
  \centering
  \includegraphics[width=\linewidth]{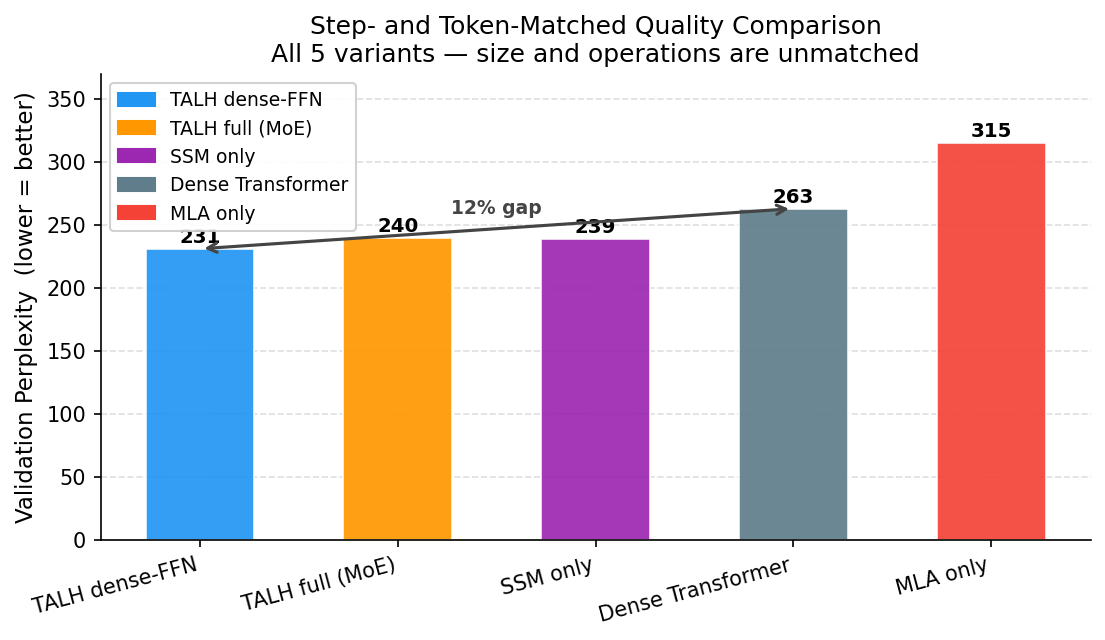}
  \caption{Step- and token-matched quality at step 4{,}000. Parameter counts,
           operations, and wall-clock training time are not matched.}
  \label{fig:phase1}
\end{figure}

Fig.~\ref{fig:phase1} and Table~\ref{tab:phase1} report validation loss and
perplexity for all five variants at step 4{,}000, the only point at which all
variants have received the same step and nominal token budget.

\begin{figure}[!t]
  \centering
  \includegraphics[width=\linewidth]{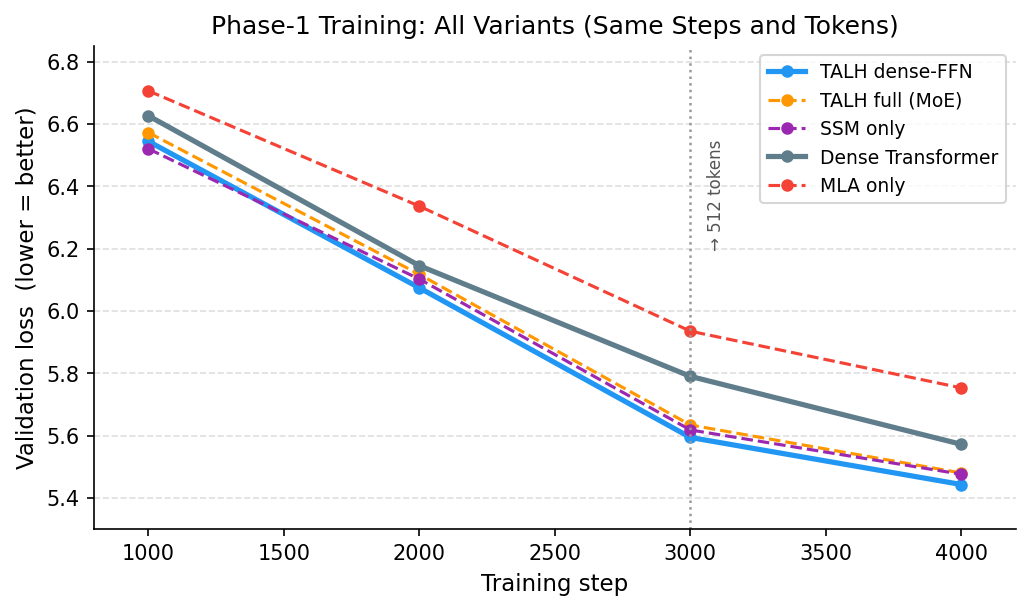}
  \caption{Validation loss curves through Phase-1 for all five variants.
           The hybrid models (\texttt{dense\_ffn}, \texttt{full}) lead
           throughout; \texttt{mla\_only} consistently trails.}
  \label{fig:phase1_curves}
\end{figure}

\begin{table}[!t]
\centering
\caption{Phase-1 results at step 4{,}000 (same steps and nominal tokens).
         PPL = $e^{\text{val\_loss}}$ at 512-token context.}
\label{tab:phase1}
\begin{tabular}{lccc}
\toprule
Variant & Val.\ loss & PPL & GPU memory \\
\midrule
\texttt{dense\_ffn}         & 5.443 & \textbf{231} & 24.01~GB \\
\texttt{ssm\_only}          & 5.476 & 239          & 26.50~GB \\
\texttt{full}               & 5.480 & 240          & 27.88~GB \\
\texttt{transformer\_dense} & 5.572 & 263          &  5.36~GB \\
\texttt{mla\_only}          & 5.753 & 315          & 16.20~GB \\
\bottomrule
\end{tabular}
\end{table}

Three observations emerge from this step- and token-matched comparison.

\textbf{Observation 1: The dense hybrid has 12\% lower measured PPL than the
baseline.}
\texttt{dense\_ffn} achieves PPL~231 against 263 for \texttt{transformer\_dense},
a 12\% reduction under the same step and nominal token budget. The hybrid has
54M more active parameters and used substantially more wall-clock time, and the
gap has not been replicated across seeds; it is therefore not an
architecture-only or statistically tested effect.

\textbf{Observation 2: Removing the recurrent branch hurts quality most.}
\texttt{ssm\_only} (PPL~239) and \texttt{full} (PPL~240) both outperform
\texttt{transformer\_dense} (PPL~263), while \texttt{mla\_only} (PPL~315) is
the worst-performing variant, worse even than the standard Transformer
baseline. Within this run and implementation, this is evidence that the custom
recurrent branch contributes more to validation loss than MLA. It does not
establish the same role decomposition for other SSM or MLA designs.

\textbf{Observation 3: The tested ternary MoE does not improve quality.}
\texttt{dense\_ffn} (PPL~231) outperforms \texttt{full} with MoE (PPL~240)
while using 3.87~GB less peak training memory. The training implementation
retains all eight experts as FP32 master weights and quantises them during the
forward pass. We did not measure routing specialisation, so the experiment does
not identify why this MoE configuration underperforms.

\subsection{Extended Training: \texttt{full} and \texttt{dense\_ffn}}
\label{sec:extended}

\begin{figure}[!t]
  \centering
  \includegraphics[width=\linewidth]{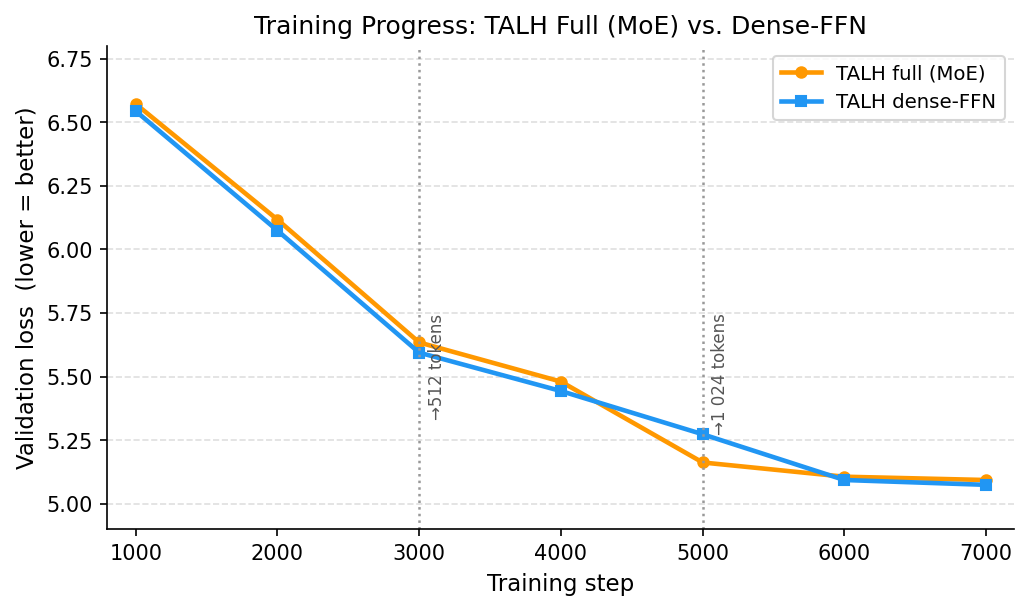}
  \caption{Extended training: \texttt{dense\_ffn} and \texttt{full} variants
           through the 1{,}024-token curriculum stage. \texttt{dense\_ffn}
           reaches the same loss as \texttt{full} approximately 1{,}000 steps
           sooner.}
  \label{fig:training}
\end{figure}

Fig.~\ref{fig:training} shows training dynamics for the two extended variants
through all three curriculum stages. Table~\ref{tab:extended_progress} reports
validation loss at key checkpoints.

\begin{table}[!t]
\centering
\caption{Extended training: validation loss at key steps. These numbers cannot
         be compared against the three variants not extended to 7K.}
\label{tab:extended_progress}
\begin{tabular}{cccc}
\toprule
Step & Context & \texttt{full} & \texttt{dense\_ffn} \\
\midrule
2{,}000 & 256  & 6.117 & 6.074 \\
4{,}000 & 512  & 5.480 & 5.443 \\
6{,}000 & 1024 & 5.106 & 5.093 \\
7{,}000 & 1024 & \textbf{5.093} & \textbf{5.074} \\
\bottomrule
\end{tabular}
\end{table}

The 1{,}024-token curriculum stage required reducing batch size from 4 to 2
(gradient accumulation doubled to 16) to avoid out-of-memory errors; training
converged cleanly. Peak memory for \texttt{full} reached 28.75~GB at this
stage vs.\ 24.88~GB for \texttt{dense\_ffn}, a 3.87~GB gap consistent with
the cost of storing all eight expert weight matrices. At step 7{,}000,
\texttt{dense\_ffn} achieves val~loss~5.074 vs.\ 5.093 for \texttt{full},
preserving the Phase~1 ordering: dense FFN matches or beats MoE at every
reported checkpoint.

\texttt{dense\_ffn} also reaches an equivalent loss to \texttt{full} about
1{,}000 steps earlier (step~6{,}000 vs.\ step~7{,}000), a 14\% reduction in
wall time for the same quality level.

\subsection{Quality vs.\ Memory Trade-off}

\begin{figure}[!t]
  \centering
  \includegraphics[width=\linewidth]{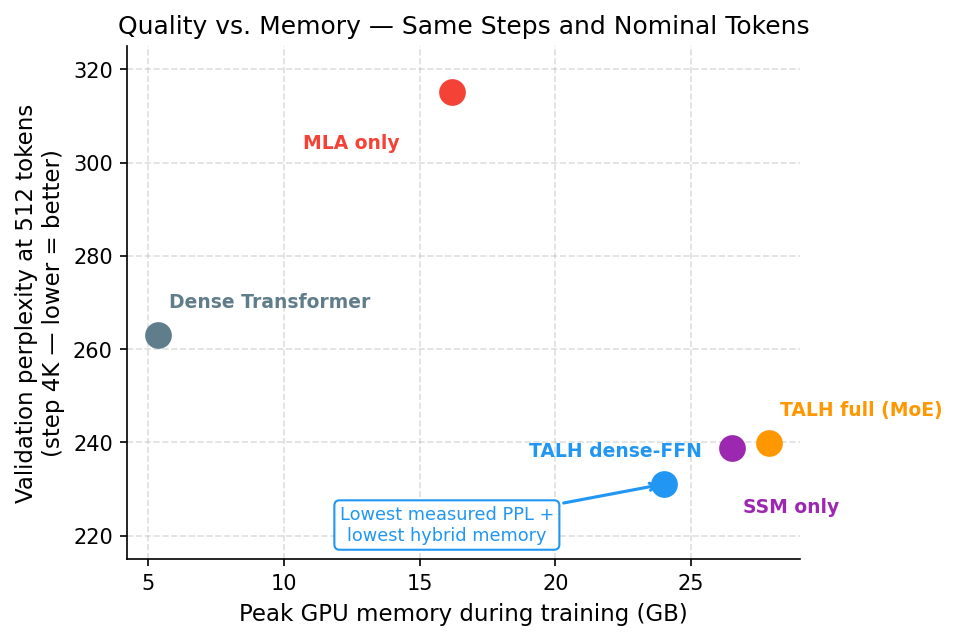}
  \caption{Step-4{,}000 quality vs.\ peak training memory. All variants use
           the same steps and nominal tokens but differ in size and operations.}
  \label{fig:pareto}
\end{figure}

Fig.~\ref{fig:pareto} plots perplexity against peak GPU training memory for
all five variants at step 4{,}000. Among the three hybrid variants,
\texttt{dense\_ffn} has the lowest measured perplexity and peak memory.
\texttt{transformer\_dense} uses far less memory (5.36~GB) but has 14\% higher
measured perplexity. The roughly 19~GB gap also reflects different active
parameter counts, optimiser state, the recurrent branch, MLA projections, and
gating; it cannot be assigned to useful architectural capacity alone.

\subsection{Local Inference Latency on Apple Silicon}

\begin{figure}[!t]
  \centering
  \includegraphics[width=\linewidth]{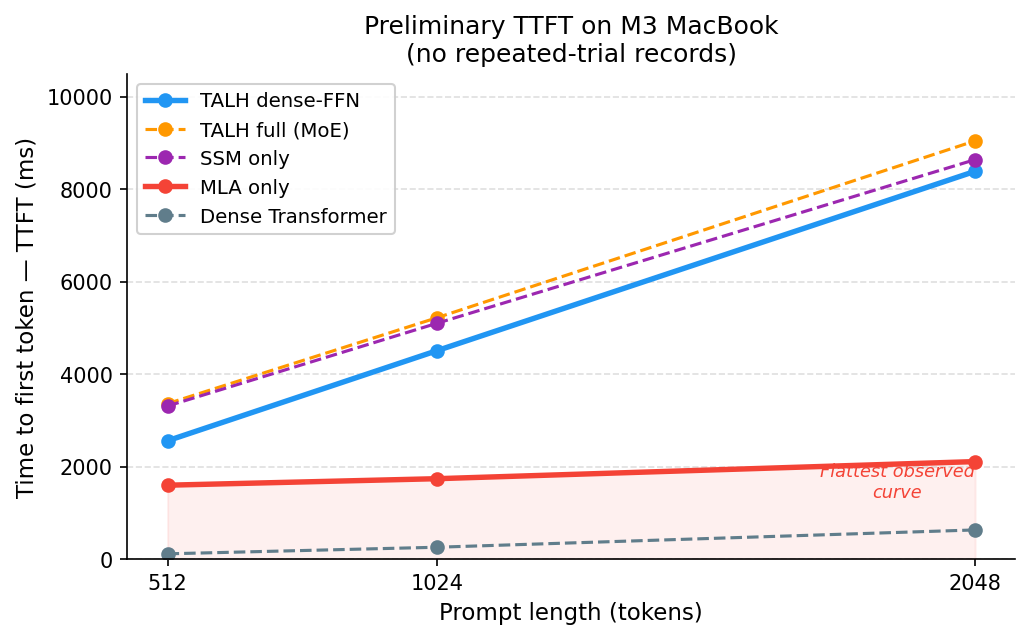}
  \caption{Preliminary single-run TTFT observations on an M3 MacBook.
           \texttt{mla\_only} has the flattest observed curve (1.32$\times$
           from 512 to 2{,}048 tokens), but no repetitions or uncertainty
           estimates are available.}
  \label{fig:ttft}
\end{figure}

Table~\ref{tab:ttft} and Fig.~\ref{fig:ttft} report TTFT on the MacBook M3
for all five variants at three prompt lengths.

\begin{table}[!t]
\centering
\caption{Time to first token (TTFT, ms) on MacBook M3, 18~GB unified memory.
         Unoptimised prototype; no quantisation, no kernel fusion.}
\label{tab:ttft}
\begin{tabular}{lccc}
\toprule
Variant & 512 tok & 1024 tok & 2048 tok \\
\midrule
\texttt{transformer\_dense} &   114~ms &    256~ms &    631~ms \\
\texttt{mla\_only}          & 1{,}599~ms & 1{,}739~ms & 2{,}110~ms \\
\texttt{dense\_ffn}         & 2{,}560~ms & 4{,}508~ms & 8{,}390~ms \\
\texttt{ssm\_only}          & 3{,}312~ms & 5{,}100~ms & 8{,}643~ms \\
\texttt{full}               & 3{,}364~ms & 5{,}221~ms & 9{,}050~ms \\
\bottomrule
\end{tabular}
\end{table}

\textbf{Observation 4: MLA-only has the flattest observed TTFT curve.}
\texttt{mla\_only} TTFT grows by only $1.32\times$ as context increases from
512 to 2{,}048 tokens, a fourfold increase in prompt length. All other
variants grow more over these three points: \texttt{transformer\_dense} by $5.5\times$,
\texttt{dense\_ffn} by $3.3\times$, \texttt{full} by $2.7\times$.
Because TTFT is dominated by prompt prefill, and because repetitions, variance,
and inference-memory measurements are unavailable, this pattern neither
establishes an asymptotic complexity nor isolates KV-cache compression as its
cause. It is a hypothesis for a future controlled benchmark.

\textbf{Absolute latency gap.} \texttt{transformer\_dense} is
substantially faster than any hybrid variant in absolute terms (114~ms
vs.\ 2{,}560~ms at 512 tokens). This is not primarily an architectural
property; it reflects two implementation differences. First, standard MHA
on Apple Silicon benefits from highly optimised fused attention kernels in
the MLX framework; the SSM scan and MLA up-projection/down-projection
operations are currently run as separate unfused kernels. Second,
\texttt{transformer\_dense} has fewer total operations per token (no SSM,
Production-quality SSM and MLA kernels may reduce the absolute gap; the present
numbers describe only this prototype.

\section{Discussion}

\textbf{How should the component ablation be interpreted?} The same qualitative
pattern appears across the five variants:
removing the SSM (mla\_only) produces the worst perplexity; removing MLA
(ssm\_only) leaves quality close to the full ternary-MoE hybrid. This motivates
the hypothesis that the recurrent branch contributes more to language-model
loss in this configuration. A single seed, unmatched parameter counts, and the
absence of a dense-FFN component ablation prevent a stronger causal claim.

\textbf{Why does the tested MoE underperform?} The experiment cannot distinguish
among optimisation difficulty from ternary forward weights, router behaviour,
expert capacity, and the different parameter allocation of the dense FFN. The
FP32 master weights and optimiser states for all experts remain resident during
training, explaining why theoretical packed-weight savings are not reflected
in peak training memory. Routing statistics and a dense-expert MoE control are
needed before attributing the result to specialisation.

\textbf{Relation to Hymba and Zebra-Llama.} Both Hymba~\cite{hymba2024} and
Zebra-Llama~\cite{zebra2025} report stronger quantitative results than TALH.
This is expected: Hymba trains on substantially more tokens with production
infrastructure, and Zebra-Llama fine-tunes from an already-trained base model.
TALH trains from scratch with a constrained budget and is reported here as an
ablation study, not as a state-of-the-art system.

\textbf{On the TTFT latency.} The large absolute TTFT gap between
\texttt{transformer\_dense} (114~ms) and \texttt{dense\_ffn} (2{,}560~ms) at
512 tokens should not be interpreted as evidence that the hybrid architecture
is unsuitable for edge deployment. It is evidence that \emph{the current
prototype implementation} is not yet optimised. Standard MHA is fast on MLX
because the framework has a fused, hardware-aware attention kernel; our SSM
Quantising weights could reduce memory traffic, but profiling is required to
identify the dominant bottleneck.

\section{Limitations}

This study has the following limitations, which should be considered when
interpreting results.

\textbf{Single training run.} All variants are trained once. Perplexity
differences, especially the small gap between \texttt{dense\_ffn},
\texttt{ssm\_only}, and \texttt{full} at Phase~1 (PPL 231, 239, 240), may not
be statistically robust without multi-seed replication.

\textbf{Parameter imbalance.} At 171M active parameters,
\texttt{dense\_ffn} has 54M more than \texttt{transformer\_dense} (117M).
The \texttt{full}, \texttt{ssm\_only}, and \texttt{mla\_only} variants have
approximately 217M, 203M, and 177M active parameters per token, respectively,
and store 453--493M total weights. The 12\% quality advantage over the
Transformer baseline is therefore not purely architectural.

\textbf{FFN width.} All variants use $d_\text{ff}=1{,}600$ (2$\times$
$d_\text{model}$). A standard 4$\times$ ratio would give $d_\text{ff}=3{,}200$
and increase \texttt{transformer\_dense} to $\approx$163M parameters. The
quality gap would likely be smaller under a standard FFN configuration.

\textbf{Matching and training cost.} Phase~1 matches steps and nominal tokens,
not FLOPs, parameters, or wall-clock time. From the released logs, 4{,}000
steps took approximately 0.58~h for \texttt{transformer\_dense}, 4.34~h for
\texttt{mla\_only}, and 19.3--20.5~h for the recurrent hybrids. Phase~2 results
for \texttt{dense\_ffn} and \texttt{full} also cannot be compared against the
three variants that were not extended.

\textbf{Evaluation-stream provenance.} The local FineWeb path constructs
validation by skipping 10{,}000 documents in the same two-shard source used for
training. The archived artifacts do not record document identifiers consumed
by each run, so train--evaluation overlap cannot be ruled out. This limits the
absolute perplexity values, although all variants use the same procedure.

\textbf{Perplexity only.} No downstream task evaluation (e.g., MMLU, ARC,
HellaSwag) is reported. Perplexity on FineWeb is a language-modelling metric
and does not directly measure reasoning or instruction-following ability.

\textbf{Inference benchmarking is incomplete.} TTFT is reported but decode
throughput (tokens/s), peak inference memory, and power consumption are not.
The prototype stack uses no quantisation, kernel fusion, or prefix caching.
Raw repetitions, software versions, and paper-scale checkpoints are not
archived, so measurement variance and exact reproduction cannot be assessed.
Comparison against production-optimised inference frameworks would be inappropriate.

\section{Conclusion}

This exploratory ablation yields two hypotheses about the tested implementation.
First, the custom recurrent branch appears more important to validation loss:
removing it produces the weakest perplexity in the study, while removing MLA
leaves quality close to the full ternary-MoE hybrid. Second, MLA-only has the
flattest preliminary TTFT curve across the three measured prompt lengths.
Neither observation establishes a general role decomposition across MLA--SSM
architectures.

On the inference side, the MLA-only variant's recorded TTFT grows by
$1.32\times$ for a fourfold context increase (512 to 2{,}048 tokens), while the
other variants grow $2.7$--$5.5\times$. Without repeated trials, memory
measurements, or a prefill/decode decomposition, this result should be treated
as preliminary rather than as confirmation of complexity or causality.

The dense feed-forward hybrid obtains lower loss than the tested ternary-MoE
variant while using 3.87~GB less peak training memory. This is a useful negative
result for that configuration, not evidence of a universal MoE scale threshold.
The clearest next steps are disjoint-data and multi-seed replication,
parameter/FLOP-matched controls, routing diagnostics, and repeated prefill,
decode, and memory benchmarks with architecture-aware checkpoint conversion.


\end{document}